\documentclass[letterpaper, 10 pt, conference]{ieeeconf}  % Comment this line out if you need a4paper

\IEEEoverridecommandlockouts                              % This command is only needed if 
\usepackage{cite}
\usepackage{array}
\usepackage{graphics} % for pdf, bitmapped graphics files
\usepackage{multirow}
\usepackage{flushend}

\usepackage{url}
\usepackage{dsfont}
\usepackage{algorithmicx}
\usepackage{algpseudocode}
\usepackage{graphicx,caption}
\usepackage{amsmath,amssymb,stmaryrd,mathtools}
\usepackage{algorithm}
\usepackage{booktabs}
\usepackage{hyperref}
\usepackage{acronym}
\usepackage{color,xcolor}
\usepackage{verbatim}
\usepackage{subcaption}

\graphicspath{{./figures/}}

\newcommand{\ignore}[1]{}
\DeclareFontFamily{U}{MnSymbolC}{}
\DeclareSymbolFont{MnSyC}{U}{MnSymbolC}{m}{n}
\DeclareFontShape{U}{MnSymbolC}{m}{n}{
    <-6>  MnSymbolC5
   <6-7>  MnSymbolC6
   <7-8>  MnSymbolC7
   <8-9>  MnSymbolC8
   <9-10> MnSymbolC9
  <10-12> MnSymbolC10
  <12->   MnSymbolC12%
}{}
\DeclareMathSymbol{\powerset}{\mathord}{MnSyC}{180}

\definecolor{orange}{rgb}{1, .36, .08}
\definecolor{darkmagenta}{rgb}{0.698,0,0.698}

\definecolor{vg_edit_color}{rgb}{0, 0.0, 1.0}

\usepackage{todonotes}
\usepackage{soul}
\definecolor{smoothgreen}{rgb}{0.7,1,0.7}
\sethlcolor{smoothgreen}

\title{\LARGE \bf People as Sensors: Imputing Maps from Human Actions                                      }

\author{Oladapo Afolabi*, Katherine Driggs-Campbell*, Roy Dong, Mykel J. Kochenderfer, and S. Shankar Sastry
\thanks{* These authors contributed equally to this work.
}% <-this % stops a space
\thanks{
This material is based upon work supported by the Office of Naval Research (ONR) under grant number 31701-23800-44--EHS1S.
Toyota Research Institute (``TRI")  provided funds to assist the authors with their research but this article solely reflects the opinions and conclusions of its authors and not TRI or any other Toyota entity.
}% <-this % stops a space
\thanks{O. Afolabi, R. Dong, and S.S. Sastry are with the Department of Electrical Engineering and Computer Sciences, University of California at Berkeley, Berkeley, CA USA (e-mail: \{oafolabi,roydong,sastry\}@eecs.berkeley.edu).}%
\thanks{K. Driggs-Campbell and M.J. Kochenderfer are with the Aeronautics and Astronautics Department, Stanford University, Stanford, CA, USA (e-mail: \{krdc,mykel\}@stanford.edu).}%
}

\begin{document}

\maketitle

%%%%%%%%%%%%%%%%%%%%%%%%%%%%%%%%%%%%%%%%%%%%%%%%%%%%%%%%%%%%%%%%%%%%%%%%%%%%%%%%
\begin{abstract}
Despite growing attention in autonomy, there are still many open problems, including how autonomous vehicles will interact and communicate with other agents, such as human drivers and pedestrians. 
Unlike most approaches that focus on pedestrian detection and planning for collision avoidance, this paper considers modeling the interaction between human drivers and pedestrians and how it might influence map estimation, as a proxy for detection.
We take a mapping inspired approach and incorporate people as sensors into mapping frameworks.
By taking advantage of other agents' actions, we demonstrate how we can impute portions of the map that would otherwise be occluded.
We evaluate our framework in human driving experiments and on real-world data, using occupancy grids and landmark-based mapping approaches. 
Our approach significantly improves overall environment awareness and out-performs standard mapping techniques.
\end{abstract}

%%%%%%%%%%%%%%%%%%%%%%%%%%%%%%%%%%%%%%%%%%%%%%%%%%%%%%%%%%%%%%%%%%%%%%%%%%%%%%%%

\section{Introduction}
Despite growing attention in autonomous driving, there are still many open problems, including how autonomous vehicles will interact and communicate with human agents~\cite{driggscampbell2017}.
These concerns are particularly important when considering vulnerable users like pedestrians~\cite{chen2017evaluation}.
Although there has been some work in vehicle control in the presence of pedestrians, the majority of research has been focused on improving perception for pedestrian detection~\cite{bandyopadhyay2013intention,gelbal2017elastic,dollar2012pedestrian}.

\begin{figure}[!t]
    \centering
    \includegraphics[width=\columnwidth]{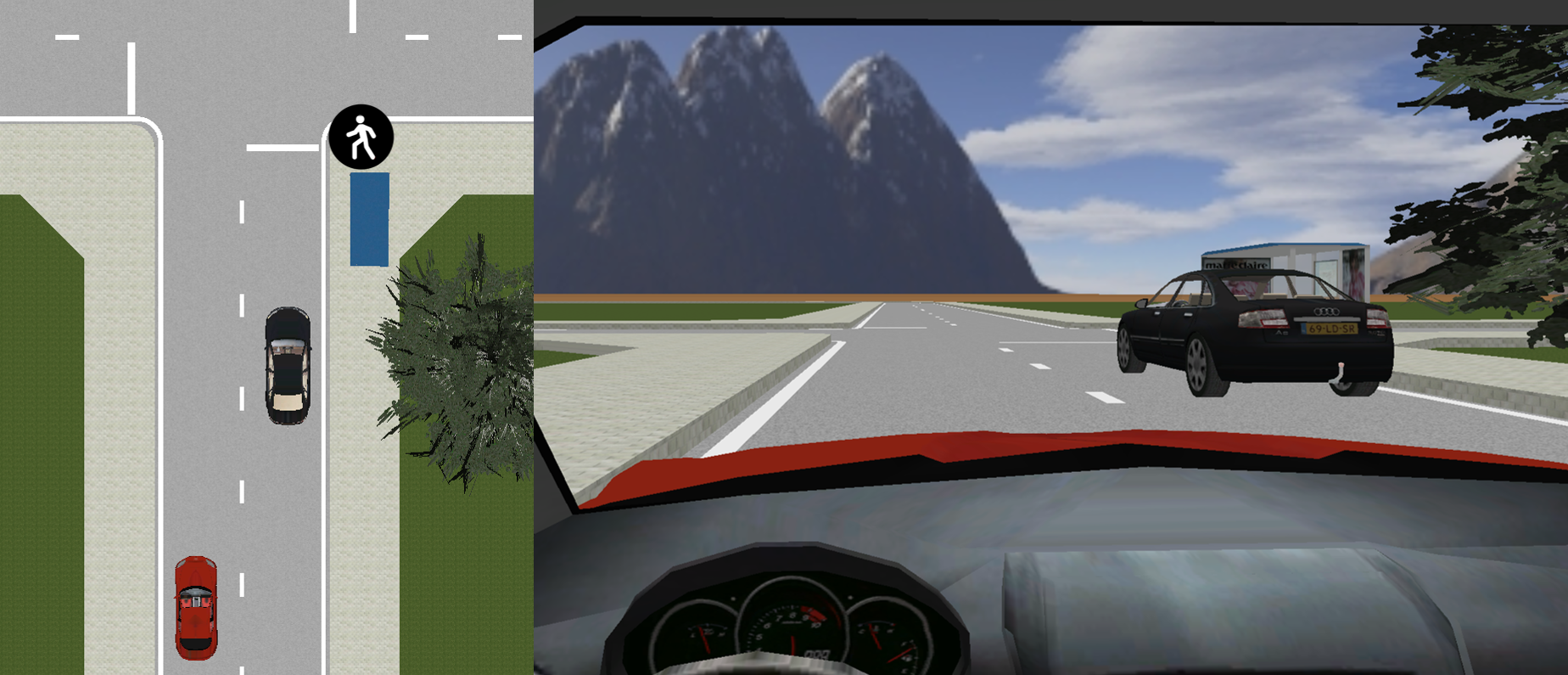}
    \caption{\small Motivating example of occluded pedestrian. \emph{(Left)} Topview showing the red ego vehicle, the black car causing occlusion, and the hidden pedestrian. We model the actions of the black car as sensor inputs. \emph{(Right)} Viewpoint from the red car (ego vehicle), showing that the pedestrian is occluded.}
    \label{fig:overview}
\end{figure}

While detection is important for a complete autonomous system, this paper considers a specific scenario concerning the interaction between pedestrians and drivers, and examines how that interaction might influence map estimation, as a proxy for detection.
Such a scenario is shown in Figure \ref{fig:overview}.  
In this scene, a pedestrian may be starting to cross the street. From the perspective of the red car, the human is occluded. We examine how to take advantage of other agent's actions to infer the presence of a pedestrian despite occlusion.
We present an approach based on a map estimation framework.

Mapping in mobile robotics refers to the process of representing an agent's environment. Based on this representation of the environment, the agent can make intelligent decisions on how to behave in and interact with the environment. In our scenario, the agent is the ego vehicle.

One common representation of the environment is the occupancy grid map. The occupancy grid map represents the environment as a grid of cells whose occupancy is modeled by independent binary random variables \cite{elfes1989}. 
The occupancy grid map allows for tractability in representing large environments with considerable amount of detail and provides a starting point for more advanced representations. 
Another type of map representation is the sparse feature-based landmark map, which only represents key objects in the environment \cite{guivant2002}. There are many more representations, many of which are more detailed (e.g. point-clouds and textured meshes), yet require more computational resources \cite{thrun2002robotic}. 
The particular choice of representation is dictated by the environment, computational resources, and how the map representation will be used to make decisions. 

Common to all these representations is the need for sensor modeling. Sensor models are regularly derived from physical properties of how the sensor in question works.
For example, the pinhole model is used for visual cameras and beam models for LIDAR and ultrasound sensors~\cite{thrun2005probabilistic}.

In this work, we exploit the fact that, aside from the physical interaction of energy (e.g. light and sound) with the environment, the actions of other intelligent agents also give us useful information about the environment. 
As such, we derive a data-driven behavioral model for agents in the environment and incorporate these people as sensors.

Behavioral driver modeling is an active area of research in many different applications, ranging from driver assistance systems to improved interaction for autonomy~\cite{driggscampbell2017,shia2014,doshi2011tactical}.
In~\cite{driggscampbell2015}, the mapping of the environment state was learned with respect to the states of the surrounding vehicles.
These influences can also be learned by estimating the cost function of the driver, which can determine what actions the driver might take given some feature representations~\cite{abbeel2004apprenticeship}.
Driver models specifically considering pedestrian interactions have also been developed~\cite{salamati2013event,gueguen2015pedestrian,sun2015estimation}.

Few approaches have directly modeled the external influences of driver behavior in a manner that is amenable to improving environment estimation.
Map-based approaches require directly modeling the connection between observable states of the vehicle (i.e., that which can be observed from a nearby vehicles) and the belief over the environment.
In this work, we focus on developing a driver model that can act as a sensor for the environment.
We apply learning techniques to approximate the distribution over pre-determined driver behaviors given a map representation.
From this, we integrate the sensor model into mapping frameworks to improve our overall awareness of the environment. 
This paper presents four key contributions:
\begin{enumerate}
    \item We introduce and formalize the concept of \emph{people as sensors} for imputing maps;
    \item We conduct an experiment with human drivers in a vehicle simulator to collect data on interactions between drivers and pedestrians;  
    \item We demonstrate improved environment estimation using occupancy grids on the collected data; and
    \item We modified pedestrian motion estimation and prediction with the landmark representation of mapping, which we test on a real-world dataset.
\end{enumerate}

This paper is organized as follows.
The methodology used to integrate driver models and mapping is summarized in Section \ref{sec:methods}.
The experimental setup for the user studies is described in Section \ref{sec:exp_setup}.
Section \ref{sec:results} presents our results using our dataset.
This method is also validated using an existing real-world dataset in Section \ref{sec:realworld}.
Section \ref{sec:disc} discusses our findings and outlines future work.
\section{Methods}
\label{sec:methods}
This section provides a brief overview of mapping and the methods used to incorporate people as sensors.

\subsection{Mapping Preliminaries}
Let $\mathcal{M}$ represent the space of maps. A map $\mathbf{m} \in \mathcal{M}$ represents a possible state of the environment. In addition, let $x_{1:t}$ represent relevant information about the mobile agent up to time $t$ (e.g., pose) and $z_{1:t}$ represent information about the physical state of the environment up to time $t$, (e.g., position of other vehicles in the environment). Then, the problem of mapping can be formulated as estimating the posterior belief  at time $t$, $p_t(\mathbf{m} \mid x_{1:t}, z_{1:t})$, over the space of maps $\mathcal{M}$. 

The choice of environment representation largely determines which algorithm to use to estimate $p_t(\mathbf{m} \mid x_{1:t}, z_{1:t})$. In this paper, we have chosen to represent the environment using two different approaches, depending on the structure of the data.
We examine applying the people as sensors framework to (1) mapping the world using occupancy grids and (2) using a collection of sparse landmarks in the environment.

\subsubsection{Occupancy Grid Maps}
When the environment is represented using an occupancy grid, the world is a set of binary random variables arranged in grids. Each random variable indicates whether or not its corresponding grid cell is occupied. 
Therefore, each map $\mathbf{m}$ is a realization of a set of binary random variables. If we denote the value of the grid cell with index $i$ as $m_i$, then $\mathbf{m} = \{m_i\}_{i=1:n}$, where $n$ is the number of grid cells used to represent the world.

Unfortunately, this choice of representation results in a space of maps that grows exponentially with the number of cells. However, most mapping algorithms make a further assumption of statistical independence between each binary random variable. Due to this assumption, one may compute the posterior belief over the space of maps $\mathcal{M}$ as:
% product of M 
\begin{equation}
p_t(\mathbf{m}\mid x_{1:t}, z_{1:t}) = \prod_{i=1}^{n} p_t(m_{i}\mid x_{1:t}, z_{1:t}),
\label{eq:slam}
\end{equation}
leaving us to focus on the simpler and more tractable task of estimating $p_t(m_{i}\mid x_{1:t}, z_{1:t})$. 
Further, we make the simplifying assumption that the state of the world at any time $t$ only depends on data obtained at time $t$. This is a reasonable assumption, given rich enough sensor and mobile agent information at time $t$. As such, we may write that:
 
\begin{equation}\label{eq:ind}
p_t(\mathbf{m}\mid x_{1:t}, z_{1:t}) = \prod_{i=1}^{n} p_t(m_{i}\mid x_{t}, z_{t}).
\end{equation}

To compute $p_t(m_{i}\mid x_{t}, z_{t})$, we make use of the mapping algorithm presented by Thrun et al., \cite{thrun2005probabilistic}. Occupancy grids are typically used for mapping in static environments. 
The application of our work focuses on non-static environments, including moving vehicles and pedestrians. 
Consequently, we modify the traditional mapping algorithm by removing the time dependence across maps, thus taking a one shot approach with no prior knowledge of the environment. However, if some domain specific knowledge about the dynamics of the environment is known, it can be incorporated into  $p_t(m_{i}\mid x_{t}, z_{t})$ through a transition function that links the previous state to the current state.

\subsubsection{Landmark Representation}
When the environment is represented as a collection of sparse landmarks, the world can be viewed as a collection of salient points in the environment (e.g., people, vehicles, key buildings and natural objects). 
These salient points are termed landmarks, and the mapping task is to estimate the state of these landmarks given data obtained from sensors. 
Typically, the state of most interest is the pose of these landmarks. 
This approach represents the map $\mathbf{m}$ as a collection of $k$ landmarks $\{ \xi_{i} \}_{i = 1:k}$, so that now, $\mathbf{m} =\{ \xi_{i} \}_{i = 1:k} $.

In this work, we have chosen to use pedestrians as landmarks. 
The state of interest is their position on the 2D floor plane. Concretely, $\xi_{i} \in \mathbb{R}^2$. One could make use of a Kalman filter to estimate  $p_t(\mathbf{m}\mid x_{1:t}, z_{1:t})$, but the publicly available dataset this model was tested on did not contain enough information to do this. Alternatively, since we are interested in modeling the position of the pedestrian in cases where they are occluded, we have assumed a uniform distribution for $p_t(\mathbf{m}\mid x_{1:t}, z_{1:t})$. This simple approach represents the fact that when the pedestrian is occluded, we may have no information about its possible location due to the unpredictability of human behavior.

\subsection{Integrating Humans in Mapping}
One of the main contributions of this paper is the use of human models as a source of sensor information. We argue that the actions of intelligent agents, specifically other drivers in this scenario, are a ubiquitous source of rich information that should not be ignored. However, it is also important that this information be incorporated appropriately with other sources of information, since human agents are highly uncertain and are difficult to model. 

Given data on human driver behaviors $a$, from a set $\mathcal{A}$, we may reformulate the mapping problem as estimating $p_t(\mathbf{m}\mid x_{1:t}, z_{1:t}, a_{1:t})$, where $a_{1:t}$ is a sequence of observed data on human drivers up until time $t$. We will subsequently refer to this human behavior data as an action. 

While this idea is indeed general, we restrict ourselves to the case where we only observe actions from the closest driver in front of our ego vehicle. In future work, we will extend formulation to the scenario with multiple driving agents and lanes.

Building on the formulation discussed in the previous section, we estimate $p_t(\mathbf{m}\mid x_{1:t}, z_{1:t}, a_{1:t})$ by using Bayes' rule to fuse information obtained from driver data with our map estimate obtained using (\ref{eq:slam}). Thus, we have: 
\begin{multline}
    p_t(\mathbf{m}\mid x_{1:t}, z_{1:t}, a_{1:t}) = \\
    \frac{p_t(a_{1:t}\mid \mathbf{m},x_{1:t}, z_{1:t})p_t(\mathbf{m}\mid x_{1:t}, z_{1:t})}{p_t(a_{1:t}\mid x_{1:t}, z_{1:t})} \enspace .
\end{multline}

As before, we assume that the state of the world at any time $t$ only depends on data obtained at time $t$. Though this assumption may seem restrictive, in practice, we use actions obtained at time $t$ that contain a history of observed behavior that inherently incorporate sequences of actions. 

We also assume that given a representation of the world, the behavior of the human driver does not depend on the pose of our mobile agent or our sensor information.
While this generally might not be true, in the specific context of our application, this assumption is valid due to the relative positioning of the agents.  Taking into account the recursive influences is left as future work.
Given these assumptions, what we seek to estimate is:
\begin{equation}
    p_t(\mathbf{m}\mid x_{1:t}, z_{1:t}, a_{1:t}) = \frac{p_t(a_t\mid \mathbf{m})p_t(\mathbf{m}\mid x_t, z_t)}{p_t(a_t)}
\end{equation}

In the case where the world is represented as an occupancy grid, we have used a driver model  $p_t(a_t\mid m_{i})$ that depends on each grid cell and fused the information from the driver to estimate:
\begin{equation}
p_t(m_{i}\mid x_t, z_t, a_t) = \frac{p_t(a_t\mid m_{i})p_t(m_{i}\mid x_t, z_t)}{p_t(a_t)} \enspace .
\end{equation}
We then make use of (\ref{eq:ind}) to obtain $p_t(\mathbf{m}\mid x_t, z_t, a_t)$. 
The next section explains in detail how we obtain the driver model $p_t(a_t\mid \mathbf{m})$.

\subsection{Sensor Models for Drivers}

To model the driver as a sensor, we must learn a function that takes in map data and outputs a probability distribution over actions that the driver may take.
Our proposed framework is general enough to handle both the occupancy grid and the landmark formulation, depending on the type of driving data used to learn the sensor model.

We evoke concepts from discrete choice theory, a tool from economics that aims to describe, explain, and predict choices between discrete alternatives \cite{train2009discrete}.
We assume we have a discrete set of actions that is both \textit{exclusive} and \textit{exhaustive}, meaning that the driver must pick one and only one of the defined actions.

\subsubsection{Likelihood of Actions from Occupancy Grids} 

Supposing we have a finite collection of driver actions $\mathcal{A}$ and the assumption that each cell in the grid is an independent Bernoulli random variable, we can approximate the probability of an action given the state of cell $m_i$ empirically.
For each action, we denote this empirical distribution as $\hat{p}_{N_s}(a\mid m_i)$, where $N_s$ is the total number of trials and $a$ is the action in set $\mathcal{A}$.

We employ this method on simulation dataset from which we are able to extract the relative position of the human driven vehicle to the crosswalk along with its velocity and acceleration information. We seek to learn the distribution over this data given the current map.

Given that learning the high-dimensional distribution is computationally difficult and data intensive, we make a few simplifications to the problem. Rather than learn over the space of all positions, velocities and accelerations, we instead define a finite collection of representative samples of this space observed during experiments.

To determine what these samples should be, we cluster  over the data containing the current distance between the human driven vehicle and the crosswalk, and ten evenly spaced samples of both velocity and acceleration over the last half second. We use $k$-means clustering algorithm to identify $k$ natural groupings in the data. These can be thought of as ``actionlets'' that correspond to typical sequences of driver behaviors. We then define these clusters as the set of possible actions we may observe. In this work, we make set $k = 10$ as prescribed by grid search.

In doing this, we can easily learn the probability distribution over actions given map data, giving us an approximation of $\hat{p}_{N_s}(a_{t} \mid m_i)$.

\subsubsection{Likelihood of Action from Landmarks} 
Using the landmark interpretation of the mapping problem, the sensor model of the driver must be approximated as the probability of an action given the position of the landmark obstacle.
To do this, we apply the logit model from discrete choice theory to find this mapping~\cite{train2009discrete}.

Previous work has demonstrated that this method can determine driver actions and intent with high accuracy \cite{driggscampbell2015}.
This approach employs the EM algorithm to iteratively find the optimal linear combination of features in the dataset to estimate the probability of an action given some map configuration: $p_t(a_t\mid \mathbf{m})$.

We employ this method on a real-world dataset. While this dataset does not provide access to the vehicle state information, it alternatively provides descriptions of the ego vehicle's velocity profile that are consistent with actions used in the literature \cite{kotseruba2016joint}:
\begin{enumerate}
    \item \textit{Moving Fast: } The vehicle is moving at a speed above predetermined threshold. 
    \item \textit{Moving Slow: } The vehicle is moving at a speed below predetermined threshold. 
    \item \textit{Accelerating: } The vehicle increasing its speed. 
    \item \textit{Decelerating: } The vehicle is decreasing its speed. 
    \item \textit{Stopped: } The vehicle is stopped. 
\end{enumerate}

We make use of these descriptions as actions. Given that these actions inherently take into account time (e.g., moving fast indicates a high constant velocity for a period of time) we do not consider the the sequences of actions over time and only estimate the map given the last observed action.

\section{Case 1: Occupancy Grid Formulation}
We first evaluate our conceptual framework on the map representation of occupancy grids.
In this test case, we carry out a user study to collect ground truth information about the state of the world, which is easily translated into the discretized space of occupancy grids.

\subsection{Experimental Setup}
\label{sec:exp_setup}

In order to build the driver model for mapping purposes, training, testing, and validation driving data is required.
For the scenario considered, there are few publicly available datasets that provide the quality of data required for mapping and driver modeling purposes.
Section \ref{sec:realworld} presents the formulation and results on one of these real-world datasets.

\begin{figure}[!t]
    \vspace{5pt}
    \centering
    \includegraphics[width=.9\columnwidth]{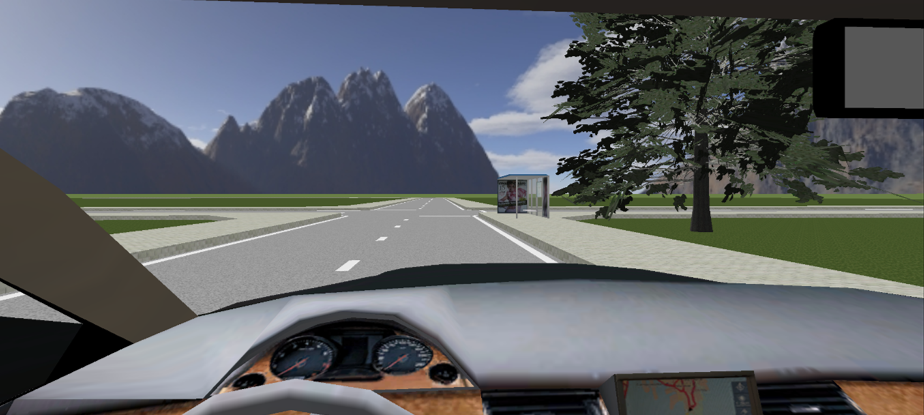}
    \caption{\small Visualization of the driver view from the experiment.  As the driver approaches the crosswalk, there is a chance a pedestrian obstacle will appear from behind the bus stop.} 
    \label{fig:exp_set}
    \vspace{-5pt}
\end{figure}

Due to lack of available data with full information about the vehicle and environment states, a new dataset was collected to study driver pedestrian interaction.
Driver data was collected using PreScan, an industry standard simulation tool that provides vehicle dynamics and customizable driving environments \cite{driggscampbell2015msthesis}.
Using a force feedback steering wheel and pedals for the subject to control the human-driven vehicle, we created various intersection scenarios in which a pedestrian might appear, as shown in Figure \ref{fig:exp_set}.

In each trial, the human-driven vehicle began approaching an intersection at an initial distance $d_0$ and speed $v_0$.
The pedestrian motion was designed to recreate typical pedestrian behaviors.
After appearing from behind an occluding obstacle at randomized velocity, the prescribed behaviors included boldly crossing the road, waiting to cross until the approaching vehicle slowed down, and just standing at the side of the road. 
To discourage anticipating the pedestrian motion, the pedestrian did not appear in half of the instances.

Five subjects each completed approximately one hour of experiments.
In each trial, the subject was asked to maintain a constant velocity between 10 and 15 mph and stay in their lane, if possible.
This resulted in 1,440 example interactions each lasting approximately 5 to 10 seconds, recorded at 30Hz.
From this, we generated a total of 281,506 maps to build our sensor models and test our mapping.
Twenty percent of this data is used to generate the learned distribution over actions.

\begin{figure}[!t]
    \centering
    \includegraphics[width=.85\columnwidth]{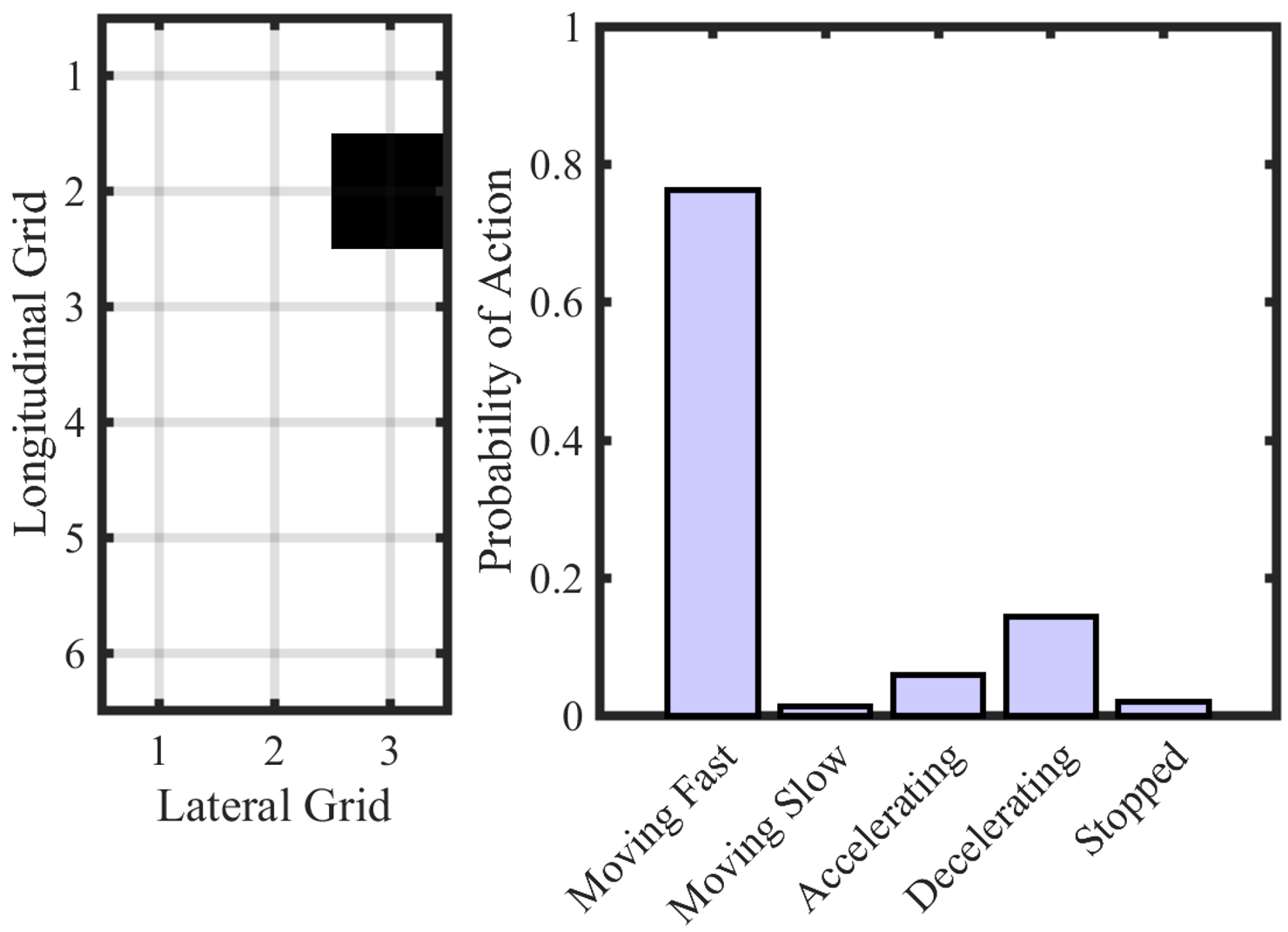}
    \caption{\small Illustrative example of input and output of the driver model. \emph{(Left)} Sample occupancy grid for the region in front of the vehicle (i.e., grid location (2,6) is directly in front of the vehicle), where an occupied space is far ahead and to the right of the vehicle at (3,2). \emph{(Right)} Probability distribution over possible semantic actions given the occupancy map on the left.}
    \label{fig:dm_example}
    \vspace{-5pt}
\end{figure}

\begin{figure*}%[!ht]
\vspace{5pt}
    \centering
    \begin{subfigure}{0.4\textwidth}
    \centering
    \includegraphics[width =\textwidth, keepaspectratio]{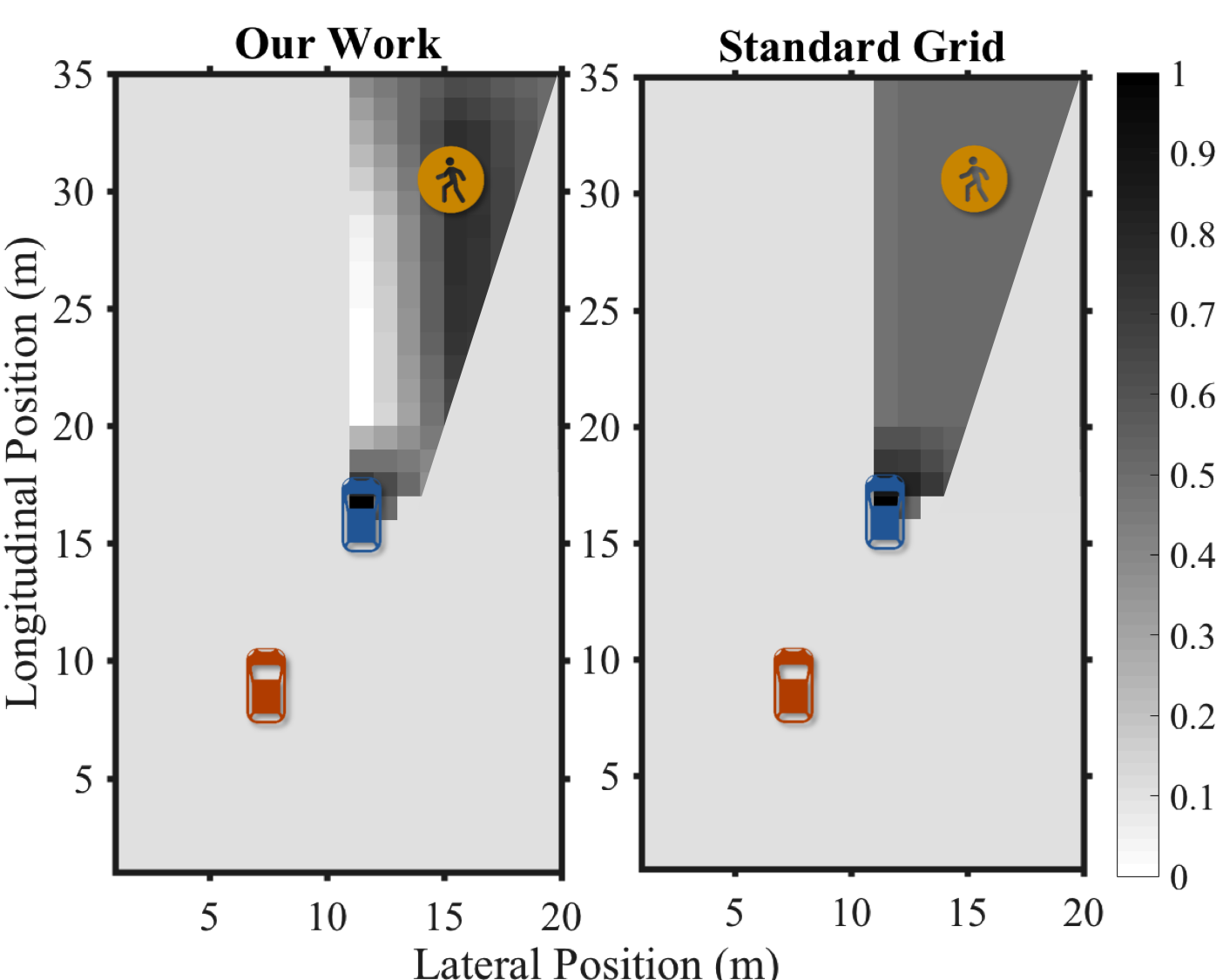}
    \end{subfigure}
    \begin{subfigure}{0.4\textwidth}
    \centering
    \includegraphics[width =\textwidth, keepaspectratio]{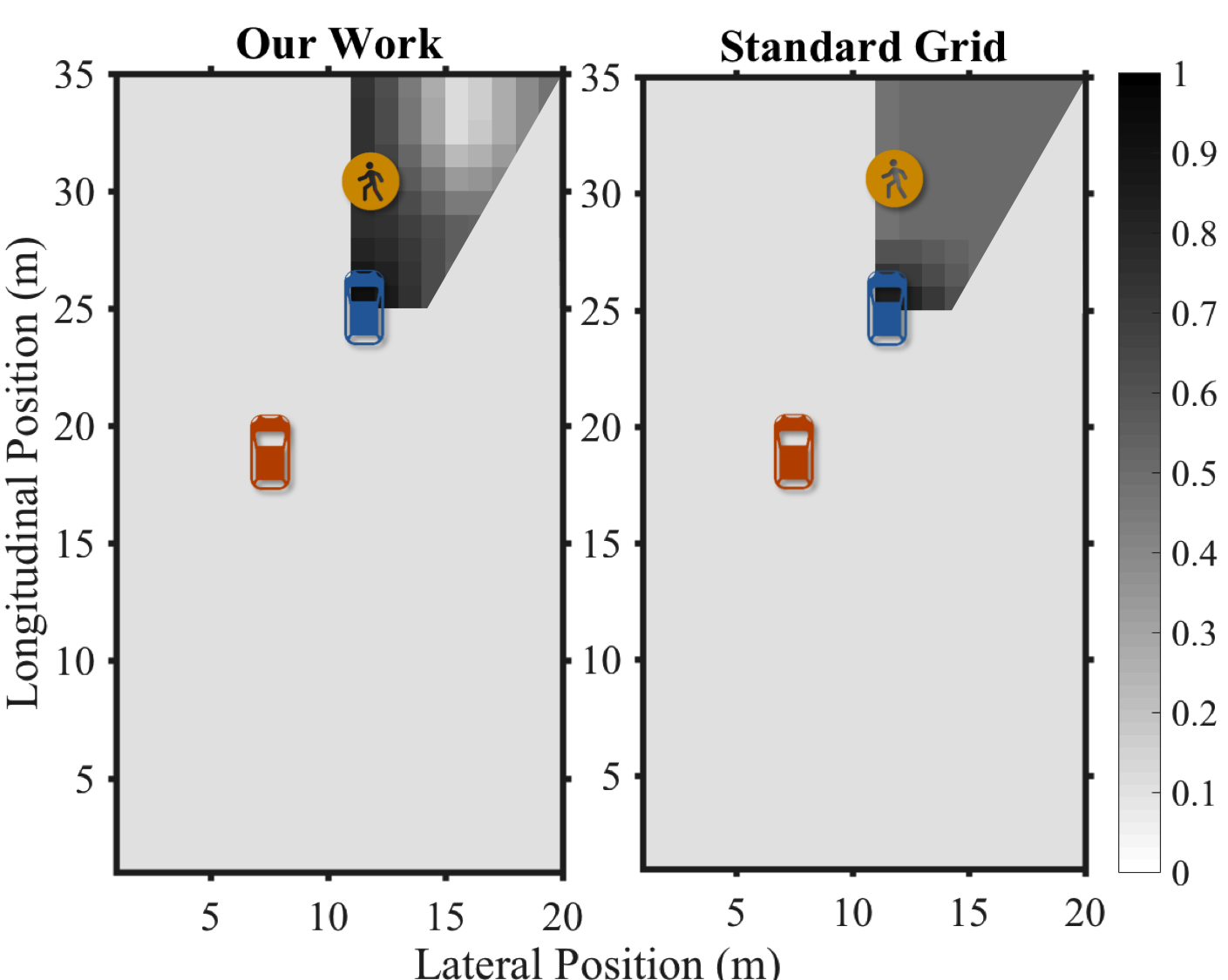}
    \end{subfigure}
    \caption{\small Two example comparisons of occupancy grids generated by our method and a standard occupancy grid algorithm. 
    Darker regions indicate greater confidence of occupancy. The orange and blue car icon represent ground truth positions of our ego vehicle and obstructing vehicle, respectively. The pedestrian icon indicates the ground truth position of the pedestrian. In the left image, the obstructing vehicle is observed increasing its velocity while in the right image the obstructing vehicle is observed slowing down. }
    \label{fig:occ_grid}
    \vspace{-5pt}
\end{figure*}

For each trial, we collected: the human driven vehicle states and inputs, and the ground truth position of the pedestrian.
Using this data, we created a ground truth occupancy for the region in front of the human driver that would be occluded for our ego vehicle.
The occlusion is determined using a simple lidar model to determine what the closest obstacles are in the 360$^\circ$ view. 
We assume only some of the vehicle states are observable from the ego vehicle (i.e., relative position and velocity, distance to crosswalk).

Using the actions defined in Section \ref{sec:methods}, we train a sensor model that maps the ground truth occupancy grid and sensor measurements to a distribution over actions; this human driver model can then be used to impute the occupancy map from observed actions.
An example occupancy grid input and the associated action distribution is shown in Figure \ref{fig:dm_example}.

To reiterate, we consider a scenario with three agents: the ego vehicle, the human driven vehicle, and the pedestrian, as visualized in Figure \ref{fig:overview}.
The ego vehicle observes the human driven vehicle that is occluding the pedestrian.
Based on the observed actions, we construct a posterior belief across possible maps.

\subsection{Evaluation Metrics}
\label{sec:results}

We tested our work on multiple scenarios from our experimental dataset. As stated previously, each scenario is composed of three agents, an ego vehicle in one lane, a human driven vehicle in the other lane causing the occlusion, and a pedestrian. The ego vehicle was set to follow a constant velocity trajectory behind the human driven vehicle in the second lane (see Fig. \ref{fig:overview}).  
To evaluate the occupancy grids generated by our approach, we made use of the Image Similarity metric.

    The Image Similarity metric, $\psi$,  is used to evaluate the similarity of an occupancy grid to an ideal or ground truth measurement \cite{birk2006merging}. This metric is computed as: 
    \begin{equation}
       \psi(A,B) = \sum_{c \in \{0,1\} } d(A,B,c) + d(B,A,c)
       \end{equation}
       where
       \begin{multline}
           d(A,B,c) = \\ \frac{1}{\#_c(A)}\sum_{A[i] = c}\underset{}{\text{min}~\left\{ ||g(i)-g(j)||_{1} : B[j] = c \right\}}
       \end{multline}
   where $A[i]$ is the occupancy value at grid cell $i$ in map $A$, $g(\cdot)$ returns the 2D coordinates of grid cell $i,j \in \{1,2,...,n\}$, $||\cdot||_1$ gives the Manhattan distance between coordinates, and $\#_c(A)$ is the number of cells in $A$ with occupancy values $c$. 
   
   To make use of $\psi$, we indicate the occupancy value of each cell by thresholding the probability $p_t(\mathbf{m}_{i} = 1|x_{1:t}, z_{1:t})$ as follows:
  \[ A[i] = 
   \begin{cases}
    1 & \text{if } p_t(m_{i} = 1|x_{1:t}, z_{1:t})\geq 0.6\\
    0 & \text{if } p_t(m_{i} = 1|x_{1:t}, z_{1:t}) < 0.6\\
   \end{cases} \enspace.
   \]

\subsection{Results}
We compare our results to a standard occupancy grid mapping algorithm  that does not incorporate information from the actions of other drivers and to ground truth measurements. We refer to the  results from the standard occupancy grid algorithm as ``Standard Grid." 

Figure~\ref{fig:occ_grid} shows  sample results based on using occupancy grids to represent the environment. The orange and blue vehicle icons indicate the ground truth positions of the ego vehicle and the human-driven vehicle respectively, while the pedestrian icon indicates the ground truth position of the pedestrian.  In these scenarios, the pedestrian is occluded from the view of the ego vehicle by the human-driven vehicle in the scene. Consequently, there is large uncertainty concerning the position of the pedestrian using the Standard Grid. Although we do not observe the pedestrian, we observe the behavior of the human-driven vehicle. By incorporating this information, our driver model is able to help us reason about the likely positions of the pedestrian. Our algorithm can reduce the uncertainty present and provide a more accurate prediction about the position of the pedestrian.

\begin{figure}[!t]
\vspace{5pt}
    \centering
    \includegraphics[width=.9\columnwidth]{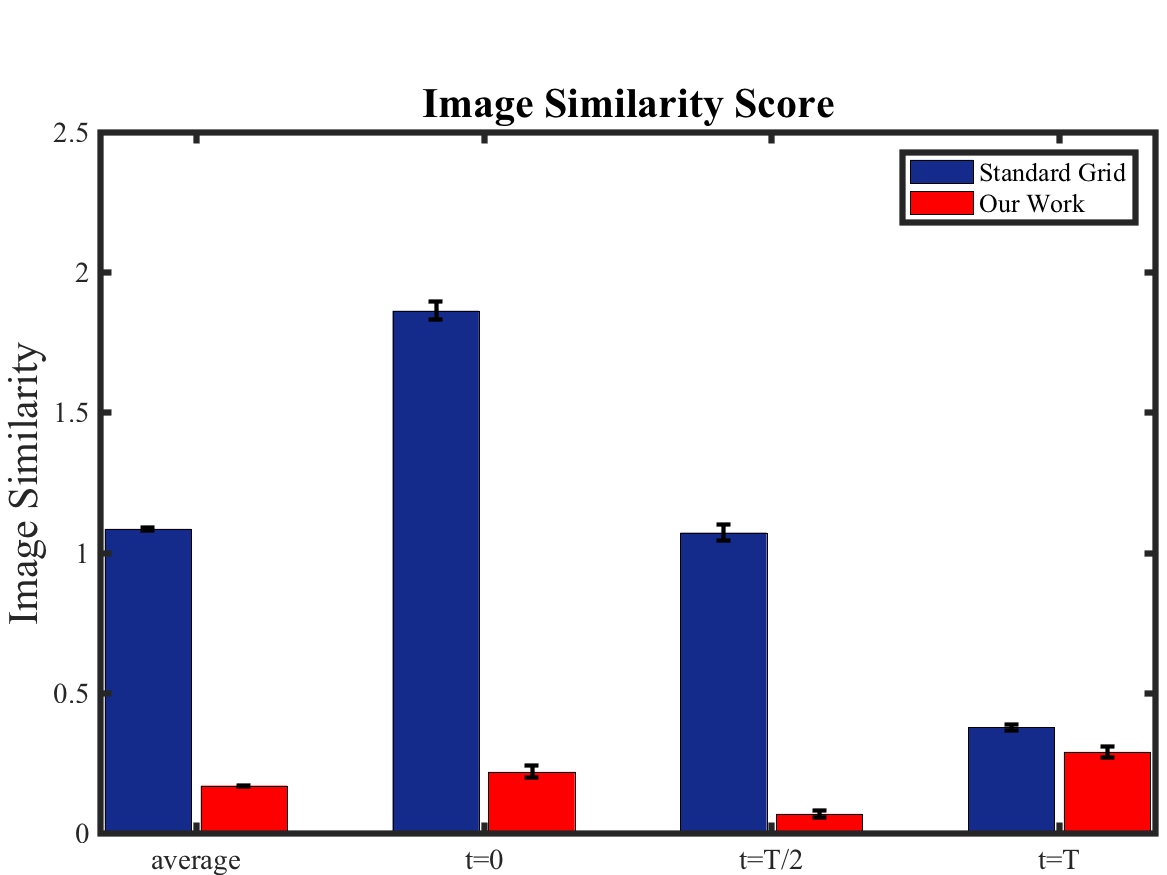}
    \caption{\small Comparison of Image Similarity scores between the occupancy grid generated by our work and the standard approaches. Lower scores imply better matching. Plots show the mean for the two methods along with the standard error. Our method exhibits significant improvement to the standard occupancy map.}
    \label{fig:birkcarpin_score}
\vspace{-10pt}
\end{figure}

Quantitatively, we applied the metric presented in the previous subsection to compare our work to the Standard Occupancy Grid. 
Table~\ref{tbl:mapping_errors} and Figure~\ref{fig:birkcarpin_score} show the average scores under the Image Similarity metric, where $ t = 0$ indicates the beginning of the trials,  $ t = T/2$ indicates the middle of each scenario, and and $T$ indicates the end of each scenario.
The mean and standard deviation of the results over time are shown in Figure ~\ref{fig:birkcarpin_score}.
Our work does significantly better than the standard occupancy grid approach.\footnote{We note that while these results show results overall drivers in the study, we also examined individual driver models.  The individual metrics exhibited similar trends to the overall metrics, except for two drivers which had significantly better performance near $t=T$.}

\begin{table}[!h]
    \centering
    \caption{\small Image Similarity Results for Occupancy Grid Approach.}
    \label{tbl:mapping_errors}
    \label{tab:map_results}
    \begin{tabular}{@{}lcccc@{}}
   
    \toprule
             & Avg. & $t = 0$ & $t = T/2$ & $t = T$ \\% & Avg. & t = 0 & t = T/2 & t = T\\
             \midrule
     Standard grid &1.085 & 1.863 & 1.071 & 0.377 \\
     Our work & \textbf{0.169} & \textbf{0.218} & \textbf{0.068} & \textbf{0.289} \\
     \bottomrule
    \end{tabular}
\end{table}

\section{Case 2: Landmarks in Real-World Dataset}
\label{sec:realworld}

We attempted to assess our framework on a more realistic scenario by testing on a real-world dataset for pedestrian interaction.
Ideally, we use a dataset with a substantial amount of pedestrian interactions, ground truth estimates of vehicle and pedestrian locations over time, and annotations of driver actions. 
Many of the public datasets do not meet these requirements. 
The Joint Attention in Autonomous Driving (JAAD) dataset centers on driver-pedestrian interaction, but provides only partial information about the state of the world through semantic action labels and approximate position estimates, making occupancy grids difficult to consider without substantial assumptions \cite{Rasouli2017}. 
Taking these restrictions into account, we modify our mapping pipeline to one that estimates  landmarks that may be occluded, and demonstrate that our framework can be applied to many different settings if context can be taken into account.

\subsection{JAAD Dataset of Pedestrian Interactions}

We use the JAAD dataset, which consists of 346 high-resolution video clips, lasting approximately 5 to 10 seconds each, that are representative of possible crosswalk scenes that often occur in urban driving.
These clips are annotated, providing labels associated with the driver and pedestrian actions as well as bounding boxes of detected pedestrians \cite{Rasouli2017}.
No vehicle state information (e.g., position, speed, acceleration) is provided with this dataset.

From the pedestrian's bounding box, we estimate the person's position relative to the vehicle camera housing to generate an approximate map for each frame.
Given the assumptions required to get this estimate, we assume a Gaussian distribution over our estimates, making this partial, noisy data more inline with the landmark philosophy.

\begin{figure}[!t]
    \vspace{5pt}
    \centering
    \includegraphics[width=.9\columnwidth]{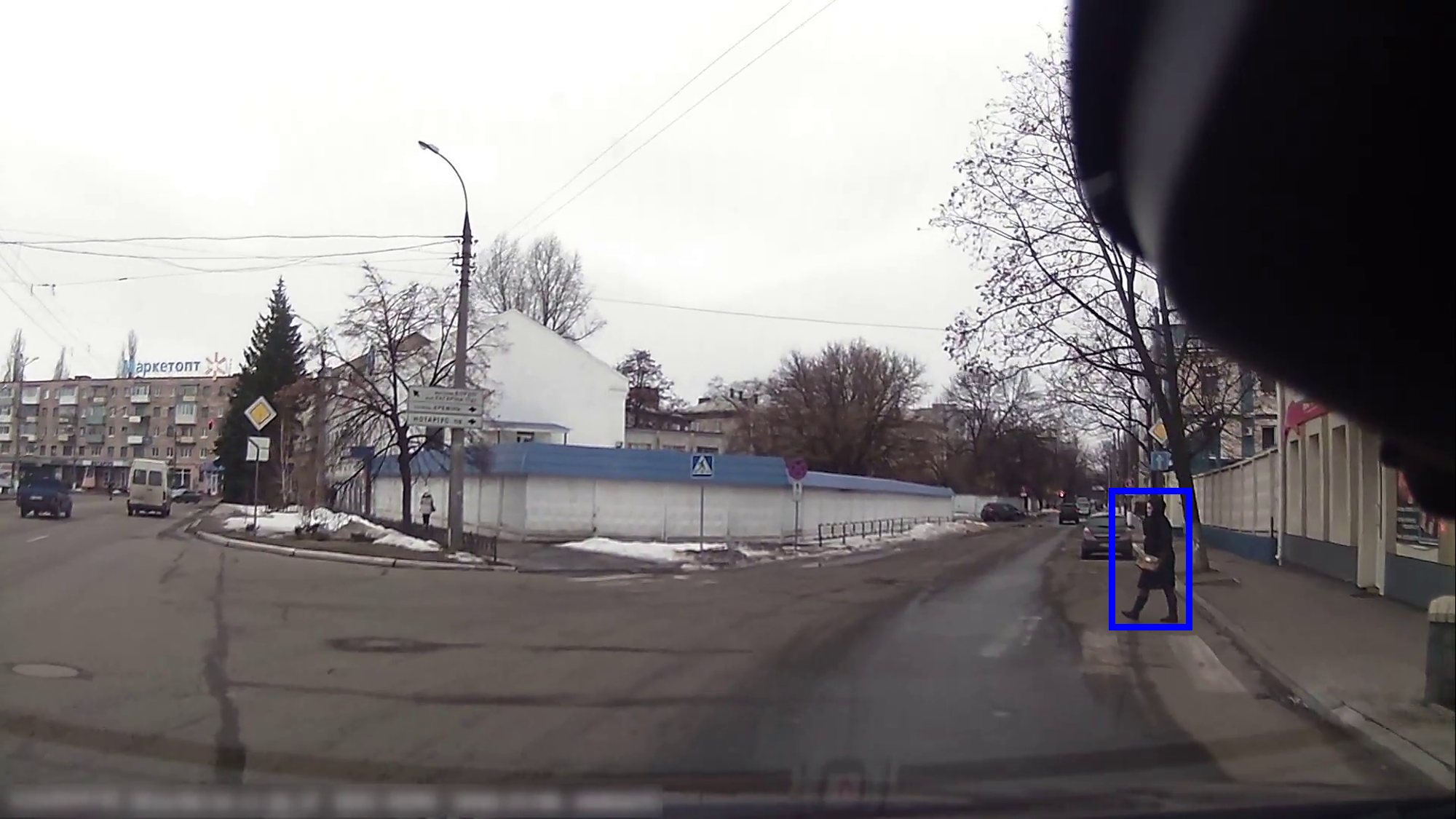}
    \caption{\small Example image from JAAD Dataset with the pedestrian in a labeled blue bounding box  \cite{Rasouli2017}.}
    \label{fig:dataset_map}
    \vspace{-10pt}
\end{figure}

From this dataset, we extract a total of 76,514 samples to train and test our model from.
The logistic regression model is trained on 20\% of the samples to find the relationship between the action and relative position.
An example image and map data are shown in Figure \ref{fig:dataset_map}.

\subsection{Results}

Using the driver model learned from the JAAD dataset, we predict the location of the pedestrian as a landmark, as described in Section \ref{sec:methods}. 
We assume an uninformed (i.e., uniform) prior over the occluded space, and show how the output of our algorithm provides a posterior distribution conditioned on the human driver actions that can improve the estimation of the pedestrian's location. 
To evaluate the improvement that the map generated by our landmarks model of the environment provides, we compare the likelihood of the pedestrian's true location in the posterior distribution to the prior distribution.

\begin{figure}[t]
    \vspace{5pt}
    \centering
    \includegraphics[width=.65\columnwidth]{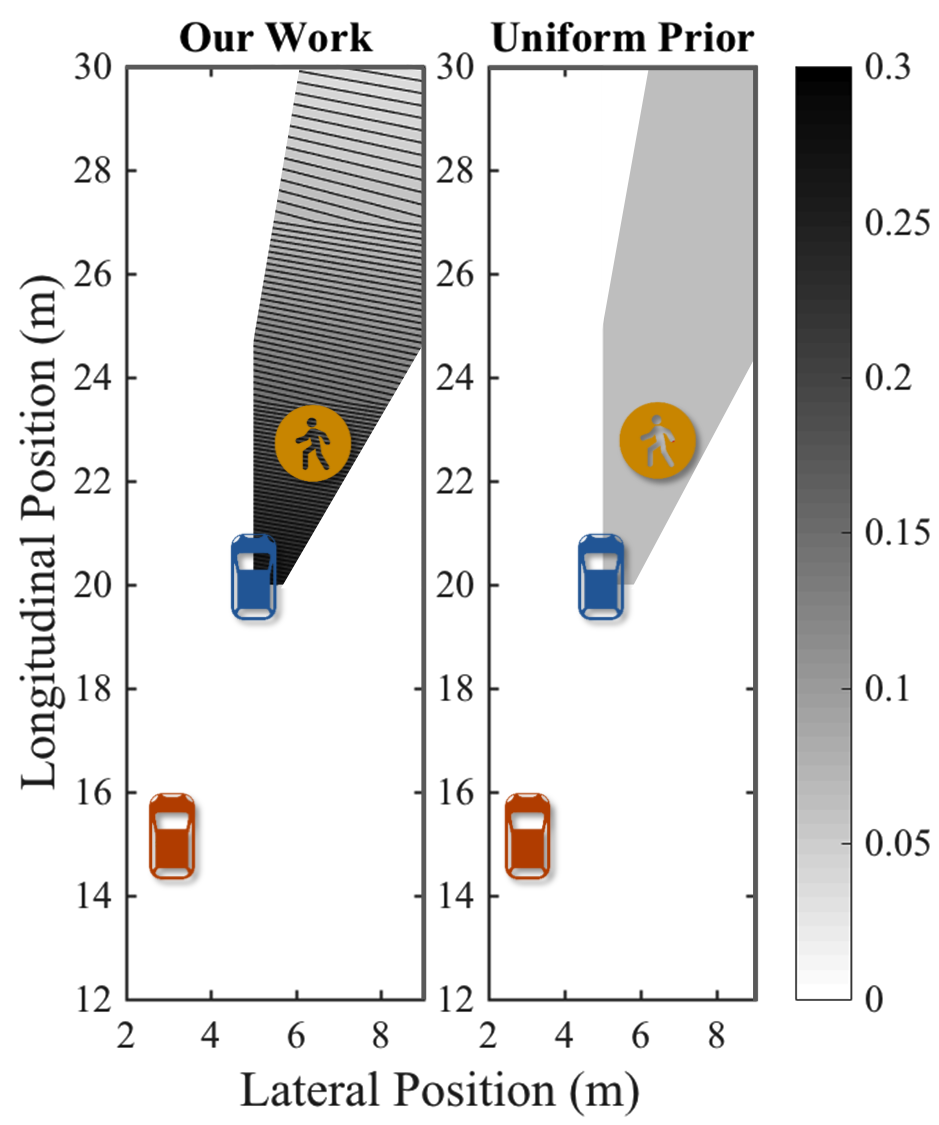}
    \caption{\small Example comparison of landmark map generated by our method versus using a uniform prior over possible locations, where the obstructing vehicle is \emph{stopped}.  White space indicates un-occluded regions.  Darker regions indicate greater belief in the pedestrian position. The orange car, blue car, and pedestrian icons represent ground truth positions of our ego vehicle, obstructing vehicle, and pedestrian, respectively.} 
    \label{fig:JAAD_results}
    \vspace{-10pt}
\end{figure}

The results of our approach compared with the uniform prior are shown in Table \ref{tab:rw_results}.
Figure~\ref{fig:JAAD_results} presents a sample output of the our work using the landmark representation and that of the uniform prior. 
The orange and blue vehicle icons represent the ground truth positions of the ego vehicle and the human-driven vehicle respectively, while the pedestrian icon represents the ground truth position of the pedestrian. 

Once again, in this scenario, the pedestrian is occluded from the view of the ego vehicle. 
The plots in the figure represent the estimated posterior density of the position of the pedestrian, with darker regions indicating higher density values. 
As shown, by incorporating the driver model learned from data, our work is able to predict the likely position of the pedestrian during occlusions. 

\begin{table}[!b]
    \centering
    \caption{\small Evaluation Metric on JAAD Dataset showing the probability of observing the pedestrian at ground truth location for each action.}
  
    \label{tab:rw_results}
    \begin{tabular}{@{}lccc@{}}
    \toprule
     \textbf{Action} & Uniform Prior & Our Work & Improvement Ratio  \\
    \midrule
     \emph{Moving Fast} & \textbf{0.064} & 0.002 & --0.963 \\
     \emph{Moving Slow} & \textbf{0.064} &0.027 & --0.569\\
     \emph{Accelerating} & 0.064 & \textbf{0.067} & 0.049 \\
     \emph{Decelerating} & 0.064 & \textbf{0.080} & 0.242 \\
     \emph{Stopped} &  0.064 & \textbf{0.257} & 3.040 \\ 
    \bottomrule
    \end{tabular}
\end{table}

Our method provides useful predictions in a majority of the actions and is most informative in safety critical situations.
The scenarios where our methodology is less informative are intuitive if we consider how drivers behave in the real-world.
When we drive and observe other drivers maintaining a constant speed, we gain little insight about occluded obstacles. We observe from the driver model derived from the JAAD dataset that the two constant velocity actions (moving fast and slow) are not informative without detailed contextual information.
Further, since we partition the dataset to only consider samples where the pedestrian might be occluded, these two actions are underrepresented relative to the other labels. 
Because of these points, our approach only exhibits improved performance on a subset of the actions. The authors are of the opinion that with ground truth position information and more evenly distributed data, significantly better improvements can be obtained as was the case in the simulated environment.

\section{Discussion}
\label{sec:disc}

By exploiting the actions of other intelligent agents, a great deal of information can be inferred about the environment.
We have presented a methodology that uses driver models as sensors to impute maps that can be used to improve planning in the face of uncertainty.
Thus, regions of the map that would otherwise be occluded can be imputed, providing an estimation of the environment's state.
We validate this concept on two different map representations and datasets, demonstrating significantly improved performance over standard mapping techniques.

While we have presented promising results on an interesting case study, there is a great deal of future work to be done.
First, given the data-driven method of this framework, there is a strong dependence on the underlying data, scenes, contexts, and semantics that are represented.
Expanding this work to more scenarios and contexts is key for making sure this works in a real-world scenario (e.g., complex intersections and crosswalks, jaywalking, general driving, etc.). 
In addition, we wish to investigate how to relax time dependence assumptions by incorporating suitable data-driven transition models. 
Finally, we believe the community would benefit from collecting datasets germane to tasks presented in this work. 
Such datasets would allow stronger comparisons and provide a better sample coverage, thus allowing for better generalization in unseen scenarios.

%\section*{Acknowledgements}
%This material is based upon work supported by the Office of Naval Research MURI Award ONR-N000141310341 and DURIP Award N000141310679.

\bibliographystyle{IEEEtran}
\bibliography{mapping_BibFile}

\clearpage

\end{document}